\documentclass[10pt,twocolumn,letterpaper]{article}

\usepackage[pagenumbers]{cvpr} % To force page numbers, e.g. for an arXiv version

\usepackage{graphicx}
\usepackage{amsmath}
\usepackage{amssymb}
\usepackage{booktabs}
\usepackage{algorithm2e}
\usepackage{float}
\usepackage[pagebackref,breaklinks,colorlinks]{hyperref}

\usepackage[capitalize]{cleveref}
\crefname{section}{Sec.}{Secs.}
\Crefname{section}{Section}{Sections}
\Crefname{table}{Table}{Tables}
\crefname{table}{Tab.}{Tabs.}
\usepackage{booktabs}
\usepackage{multirow}
\usepackage{graphicx}
\def\cvprPaperID{*****} % *** Enter the CVPR Paper ID here
\def\confName{CVPR}
\def\confYear{2022}

\begin{document}

\title{Generalist Vision Foundation Models for Medical Imaging: A Case Study of Segment Anything Model on Zero-Shot Medical Segmentation}

\author{Peilun Shi$^{\dagger}$ \and Jianing Qiu$^{\dagger, \ddagger}$ \and Sai Mu Dalike Abaxi$^{\dagger}$ 
\and Hao Wei$^{\dagger}$ \and Frank P.-W. Lo$^{\ddagger}$ \\ 
$^{\dagger}$The Chinese University of Hong Kong \\
{\tt\small \{peilunshi, tariqabaxi, wyuan\}@cuhk.edu.hk} \\
{\tt\small haowei@link.cuhk.edu.hk} 
\and Wu Yuan$^{\dagger, }$\thanks{Corresponding author} \\ 
$^{\ddagger}$Imperial College London\\ 
{\tt\small \{jianing.qiu17, po.lo15\}@imperial.ac.uk} 
}
\maketitle

%%%%%%%%% ABSTRACT
% Abstract (Do not insert blank lines, i.e. \\) 
\begin{abstract}
In this paper, we examine the recent Segment Anything Model (SAM) on medical images, and report both quantitative and qualitative zero-shot segmentation results on nine medical image segmentation benchmarks, covering various imaging modalities, such as optical coherence tomography (OCT), magnetic resonance imaging (MRI), and computed tomography (CT), as well as different applications including dermatology, ophthalmology, and radiology. Those benchmarks are representative and commonly used in model development. Our experimental results indicate that while SAM presents remarkable segmentation performance on images from the general domain, its zero-shot segmentation ability remains restricted for out-of-distribution images, e.g., medical images. In addition, SAM exhibits inconsistent zero-shot segmentation performance across different unseen medical domains. For certain structured targets, e.g., blood vessels, the zero-shot segmentation of SAM completely failed. In contrast, a simple fine-tuning of it with a small amount of data could lead to remarkable improvement of the segmentation quality, showing the great potential and feasibility of using fine-tuned SAM to achieve accurate medical image segmentation for a precision diagnostics. Our study indicates the versatility of generalist vision foundation models on medical imaging, and their great potential to achieve desired performance through fine-turning and eventually address the challenges associated with accessing large and diverse medical datasets in support of clinical diagnostics.
\end{abstract}

%%%%%%%%% BODY TEXT
\section{Introduction}\label{sec:introduction}

Recently, large AI models (LAMs) have been actively researched as they manifest impressive performance on various downstream tasks and offer a foundation to advance and foster future research in manifold AI areas, such as computer vision and natural language processing \cite{bommasani2021opportunities,mattjie2023exploring}. In medical and healthcare domains, LAMs are also transforming methodological designs and paradigms, and establishing new state-of-the-arts and breakthroughs in various sectors including medical informatics and decision-making \cite{qiu2023large}. Despite the active development, advances in medical LAMs often lag behind their counterparts in general domains. To identify current discrepancies and guide the future development of medical LAMs, we select one of these LAMs in the general domain, i.e., Segment Anything Model (SAM) \cite{kirillov2023segment}, which is a foundational vision model recently proposed for image segmentation and has shown stunning performance on tasks ranging from edge detection to instance segmentation, and thoroughly evaluate its zero-shot segmentation performance on medical images. Although there are few studies out there that tested SAM on medical imaging, they either only focus on one imaging modality, i.e., pathology \cite{deng2023segment}, or only showcase a few qualitative segmentation samples \cite{ji2023segment} without reporting quantitative results. To provide a comprehensive and objective evaluation of SAM on medical image segmentation, this work conducted extensive experiments on nine benchmarks using the zero-shot segmentation feature of SAM. The selected datasets contain a wide diversity of medical imaging modalities and organs.

Our key findings include:
\begin{enumerate}
 \item SAM demonstrated better performance on endoscopic and dermoscopic images than other medical modalities, which is conjectured as SAM was trained with a large volume of RGB image data, and endoscopic and dermoscopic images are essentially images captured by RGB cameras. Therefore, when transferred to relevant medical images, SAM can demonstrate a relatively decent and consistent performance as it is tested on general RGB images.
 \item SAM failed to carry out zero-shot segmentation tasks on images that have continuous branching structures, such as blood vessels. Interestingly enough, when tested on images of tree branches, we found SAM was actually also unable to segment them in a zero-shot manner. 
 
 % But an uncomplicated fine-tuning process with a small quantity of training data could significantly enhance the quality of segmentation.
 % \item SAM were better at segmenting objects of a closed region than objects of an open region (apart from the blood vessels, we found that the zero-shot segmentation capability of SAM also did not work well for annular objects such as optic discs), based on a provided prompt or bounding box.
 % which implies that while SAM can effectively segment areas where objects are present, it is not capable of accurately segmenting annular objects such as optic discs in the medical domain.
 
 \item Compared to models specially designed for medical imaging, the zero-shot segmentation capability of SAM on medical images are decent but often inferior to those domain-specific models. Our experiments reveal that the Dice coefficients of SAM on medical benchmarks were generally lower by 0.1--0.4 compared to previous state-of-the-art (SOTA) models in medical image segmentation. In the worst case, the Dice score of the zero-shot SAM is even lower than the SOTA by 0.65.
 
 % \item SAM had varying performance across different unseen domains, which could be attributed to differences in image appearances resulting from different imaging devices or scan protocols. 
 % \item Unlike self-supervised large language models such as GPT-4 \cite{gpt4}, SAM employs supervised training, which makes it difficult to double the amount of training data. parameters of SAM are not as numerous as those in larger language models, and currently, no higher-level discrimination or discriminatory capabilities have been observed in this model.

 \item Preliminary experiments of fine-tuning SAM were conducted. A simple fine-tuning of SAM on small amount of retinal vessel data led to impressive improvements of the segmentation quality, implying the great potential of SAM on medical image segmentation by fine-tuning. {Our code is available by {accessing} %MDPI: please add accessed date for link.
 \url{https://github.com/hwei-hw/Generalist\_Vision\_Foundation\_Models\_for\_Medical\_Imaging}.}
 
\end{enumerate}
\section{Segment Anything Model}\label{sec:sam}

SAM is a generalist vision foundation model for image segmentation, and supports a diverse range of input prompts to enhance the segmentation quality. However, SAM does not recognize the type of each individual segmented object. To facilitate comparison and evaluation, the prompt for segmenting each instance is derived from the centroid of each individual ground truth mask of that instance in our study. Upon receiving the prompt, SAM generates three potential segmentation results and provides corresponding scores. The highest-scored result is selected and compared with the ground truth for evaluation. The details of implementation are introduced in Algorithm \ref{alg:1}.

\RestyleAlgo{ruled}
\begin{algorithm}
\caption{SAM on Zero-Shot Medical Image Segmentation.}
\label{alg:1}
\footnotesize
\KwIn{Pretrained SAM Model $\theta(\cdot,\cdot)$, contour detector $CD(\cdot)$, midpoint detector $MD(\cdot)$, medical image dataset $I$ with labels of classes $C$.}
\KwOut{Segmentation mask set $M$.}
\For{$i \in I$}{

 \For{$cls \in C$}{
 $i_c \gets (label(i)==cls)$\
 
 $Contours \gets CD(i_c)$\

 Initialize image mask $m$\
 
 \For{$c \in Contours$}{
 
 $P \gets MD(c)$\
 
 $m_{outs}, scores \gets \theta(i, P)$\

 $m \gets Argmax(scores(m_{outs}))$\
 
 }
 $M$.append($m$)\
 
 }
}

\end{algorithm}

\section{Experiments}\label{sec:experiments}
Our study was implemented on a single NVIDIA RTX 3080 GPU and the official checkpoint of ViT-H SAM model was chosen to test the best performance of SAM on zero-shot medical image segmentation. After conducting tests across multiple medical imaging modalities, we observed that the segmentation outcomes were not consistently satisfactory among those modalities.

\subsection{Medical Image Segmentation Datasets}

Nine datasets (summarized in Table \ref{ALLDATA}), including Skin Lesion Analysis Toward Melanoma Detection \cite{codella2019skin,tschandl2018ham10000skin}, Drishiti-GS \cite{sivaswamy2015comprehensiveDrishti-GS}, RIM-ONE-r3 \cite{fumero2011rim}, REFUGE \cite{orlando2020refuge}, AMOS \cite{ji2022amos}, MICCAI 2017 Robotic Instrument Segmentation \cite{allan20192017robodata}, Chest X-ray \cite{candemir2013lungxray,jaeger2013automaticXRay}, Rat Colon \cite{park2017broadbandoct} and AROI \cite{melinvsvcak2021annotatedAROI} were used to examine the performance of SAM on medical image segmentation, covering a wide range of medical imaging modalities, such as OCT, MRI, and CT, as well as a diverse range of organs, including eyes, colon, spleen, kidney, gallbladder, esophagus, liver, and stomach.

\begin{table*}[]
\centering
\caption{Datasets Used for Examining the Zero-Shot Medical Image Segmentation Performance \mbox{of SAM}.}
\label{ALLDATA}
\begin{tabular}{@{}cccc@{}}
\toprule
Dataset                                        & Modality       & Details          & \multicolumn{1}{l}{Number of Test Samples} \\ \midrule
Skin Lesion Analysis Toward Melanoma Detection~\cite{codella2019skin,tschandl2018ham10000skin}  & Dermoscope     & Skin             & 259  \& 1000                                       \\
Drishiti-GS~\cite{sivaswamy2015comprehensiveDrishti-GS} RIM-ONE-r3~\cite{fumero2011rim} REFUGE~\cite{orlando2020refuge}                                           & Fundus         & Eye              & 51 \& 60 \& 160                                \\
AMOS~\cite{ji2022amos}                                           & CT             & Abdominal organs & 15361                                            \\
AMOS~\cite{ji2022amos}                                           & MRI            & Abdominal organs & 3176                                             \\
MICCAI 2017 Robotic Instrument Segmentation~\cite{allan20192017robodata}    & Endoscope      & Tissue           & 1200                                             \\
Chest X-ray~\cite{jaeger2013automaticXRay,candemir2013lungxray}                                     & X-ray          & Chest            & 704                                              \\
Rat Colon~\cite{park2017broadbandoct}                                      & Endoscopic OCT & Colon            & 130                                              \\
AROI~\cite{melinvsvcak2021annotatedAROI}                                           & SD-OCT            & Retina          & 113                                              \\ \bottomrule
\end{tabular}
\end{table*}

\subsection{Evaluation Metrics}
To assess the zero-shot segmentation capability of SAM on medical images, two quantitative metrics were employed: Dice similarity coefficient and Intersection over Union (IoU). The Dice coefficient measures the overlap between two sets of data and ranges from 0 (no overlap) to 1 (perfect overlap). Similarly, IoU computes the ratio of the intersection over the union of two sets and ranges from 0 to 1. Both metrics were then averaged across multiple samples to obtain an overall measure of the segmentation accuracy. Specifically, the Dice coefficient and IoU were calculated as

\begin{equation}
Dice = \dfrac{2 |Y \cap \hat{Y}|}{|Y| + |\hat{Y}|}
\end{equation}

\begin{equation}
IoU = \dfrac{|Y \cap \hat{Y}|}{|Y| + |\hat{Y}| - |Y \cap \hat{Y}|}
\end{equation}
where Y refers to the ground truth mask and \^{Y} denotes the predicted mask of SAM based on the prompt (i.e., the centroid of the ground truth mask). Dice and IoU are both the gold standards for evaluating the overlap of the ground truth and the predicted regions in image segmentation tasks. However, IoU is more suitable for evaluating the worst-case scenarios, which could provide a more comprehensive evaluation on the medical image segmentation. As some current methods use either one of these two evaluation metrics or both to report results, we use both Dice and IoU in our work to enable direct comparison.

\subsection{Results}

This section presents the results of zero-shot medical segmentation of SAM on eight different imaging modalities, as well as preliminary results of retinal vessel segmentation by fine-tuning SAM.

\begin{figure*}[!t]
\centerline{\includegraphics[width=\textwidth]{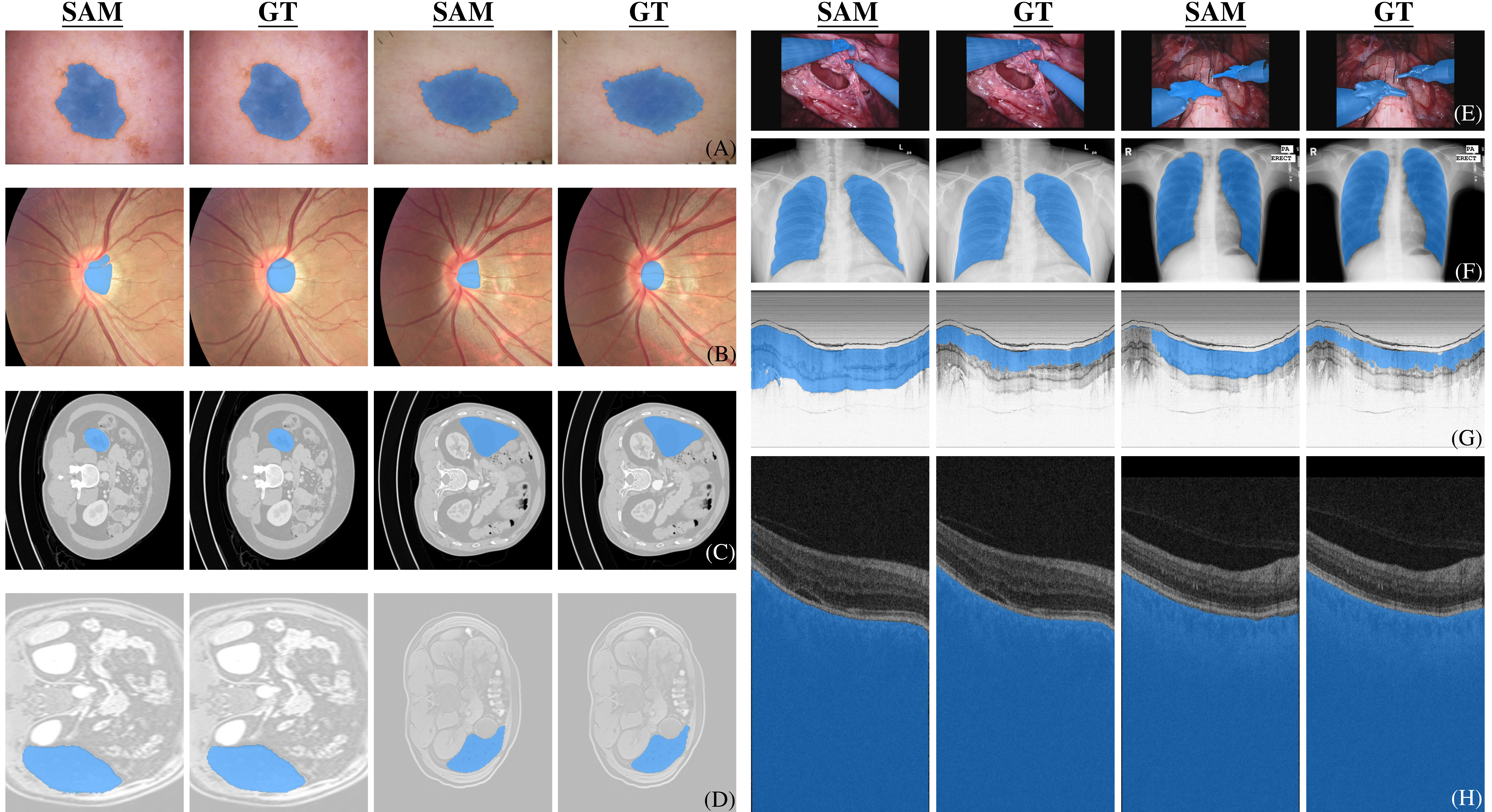}}
\caption{Successful segmentation examples of SAM. Eight distinct modalities labeled with (\textbf{A}--\textbf{H}) are included, corresponding to dermoscope, fundus, CT, MRI, RGB endoscope, X-ray, endoscopic OCT, and ophthalmic OCT. Each set of images comprises four images, containing two pairs of SAM segmentation versus corresponding ground truth (GT).}
\label{fig:good-samples}
\end{figure*}

\begin{figure*}[!t]
\centerline{\includegraphics[width=\textwidth]{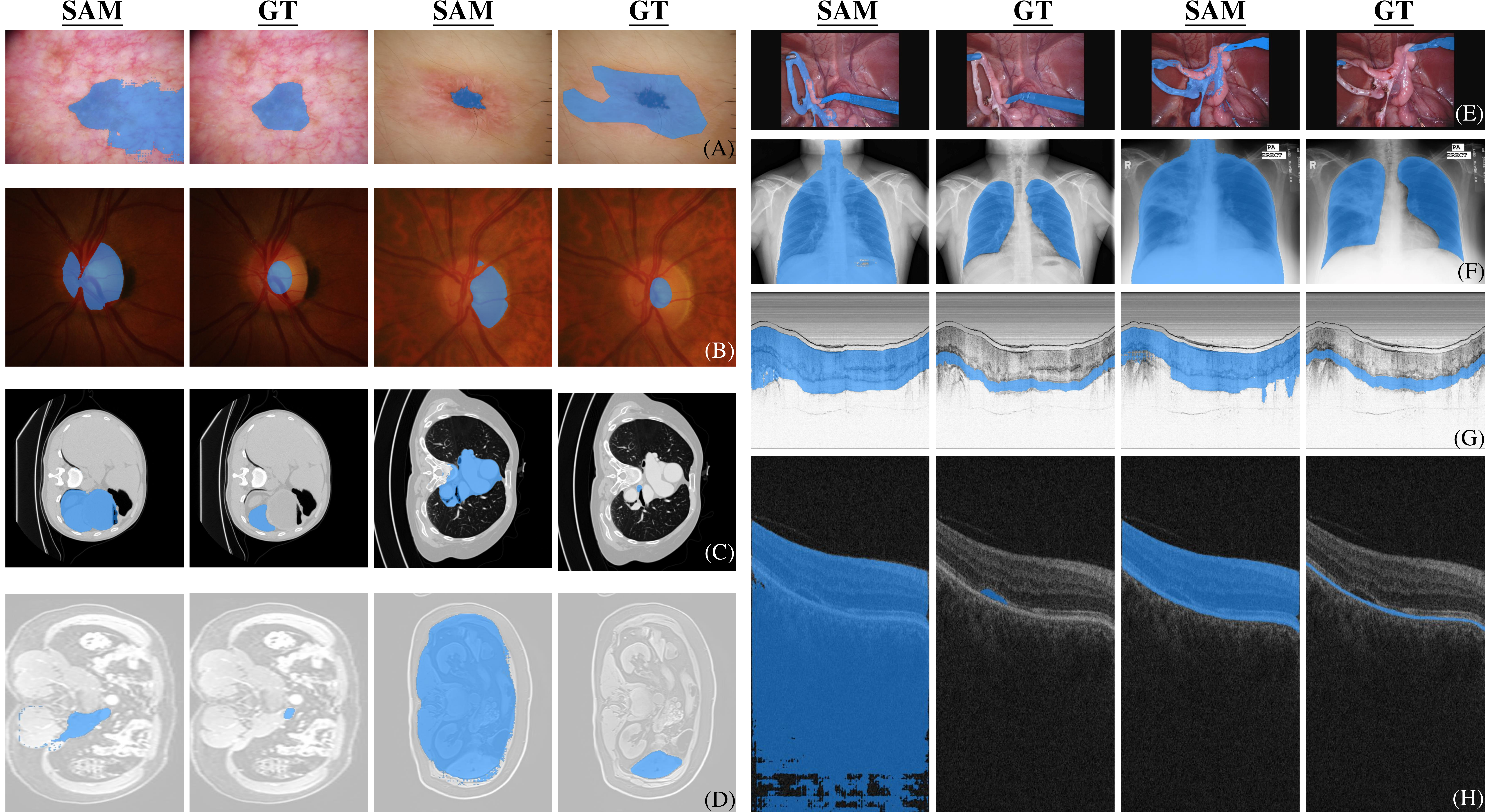}}
\caption{Failure segmentation examples of SAM. Eight distinct modalities labeled with (\textbf{A}--\textbf{H}) correspond to dermoscope, fundus, CT, MRI, RGB endoscope, X-ray, endoscopic OCT, and ophthalmic OCT. Each set of images consists of two paired SAM segmentation and ground truth (GT).}
\label{fig:bad-samples}
\end{figure*}

% \subsection{TODO}

% \begin{itemize}
%     \item \textcolor{red}{Create result table for each dataset (5 tables)}
%     \item \textcolor{red}{a figure showing good segmentation results, 4 columns, 2 examples per dataset (5 rows), using the following order for the column: sam/gt/sam/gt}
%     \item \textcolor{red}{a figure showing bad segmentation results, like above}

%     \item \textcolor{red}{a figure showing sam segmentation results that seem better than gt?? this one is optional, and we can just report one sample per dataset, or even without reporting for some datasets.}

% \end{itemize}

% \begin{enumerate}
\subsubsection{Dermoscopic Images}
Skin lesion analysis dataset is sourced from the challenge hosted by the International Skin Imaging Collaboration (ISIC) \cite{codella2019skin,tschandl2018ham10000skin}. Note that images captured by dermatoscopes with professional lighting and better magnification always present accurate details and precise contours. However, melanoma appears heterogeneously on different states of skin or diseases. Table \ref{skin} summarizes the results for comparison. From the bottom two rows of Table \ref{skin}, it is evident that for the same task, there is a significant difference in the final results due to the difference in the number of tested samples. We hypothesize that this discrepancy may be attributed to the randomness from the small number of samples. Therefore, the results from testing with larger number of samples may be more reliable. SAM (1000) refers to the performance of the model assessed on the official complete testing dataset. Additionally, as shown in Table \ref{skin}, the results of SAM are not competitive compared to existing methods. Specifically, SAM (259) denotes the outcome obtained by utilizing ten percent of randomly selected instances from the whole dataset as a testing subset, which facilitates a straightforward comparison with existing methods as using roughly ten percent of data has been a common practice in prior works. Figure \ref{fig:good-samples}A shows two samples that SAM perfectly segmented. As shown in Figure \ref{fig:good-samples}A, SAM is able to accurately segment the target when the anomalous regions are positioned at the center and exhibit a distinct morphology.

Figure \ref{fig:bad-samples}A shows two failure cases. When the target lesion locates in the erythematosus or scars, the segmentation often fails due to the similarity of features between the target and the adjacent parts or the absence of clear boundaries (or in another word, the lesion is less obtrusive or apparent).

% The quantities of both also correspond to the first row shown in Table~\ref{ALLDATA} at the same time. 
% \textcolor{red}{see my comments about in quantitative ... in review mode}

% \usepackage{booktabs}
% \usepackage{graphicx}
\begin{table}[]
\centering
\caption{Comparison with SOTA Methods on the Skin Lesion Analysis Toward Melanoma Detection Dataset.}
\label{skin}
% \resizebox{\columnwidth}{!}{%
\begin{tabular}{@{}ccc@{}}
\toprule
Method      & Dice            & IoU             \\ \midrule
 U-Net~\cite{ronneberger2015unet}     & 0.8550          & 0.7850          \\
 UNet++~\cite{zhou2018unet++}     & 0.8090          & 0.7290        \\
 CaraNet~\cite{lou2022caranet}    & 0.8700          & 0.782          \\
 PraNet~\cite{fan2020pranet}     & 0.8750          & 0.7870          \\
 TransUNet~\cite{chen2021transunet}     & 0.8800          & 0.8090          \\ TransFuse~\cite{zhang2021transfuse}     & 0.9010          & 0.840          \\
 Double-UNet~\cite{jha2020doubleu}     & 0.8962          & 0.8212          \\
 Polyp-PVT~\cite{dong2021polyp}     & 0.8962          & 0.8212          \\
 DuAT~\cite{tang2022duat}     & 0.9230          & 0.8670          \\
 Polar Res-U-Net++(SOTA)~\cite{benvcevic2021skinsota} & \textbf{0.9253} & \textbf{0.8743} \\
\bottomrule
SAM (259 samples, common practice)     & 0.6636          & 0.5566          \\

 SAM (1000 samples, official test set)    & 0.7306          & 0.6169          \\
\bottomrule
\end{tabular}%
% }
\end{table}

\subsubsection{Fundus Images}
Retinal fundus images are considered as the primary modality in ophthalmic diagnosis. Specifically, optic cup (OC) and optic disc (OD) segmentation from the fundus image is the essential part. DoFE \cite{wang2020dofe} is a benchmark for OC and OD segmentation that comprises three different fundus image datasets. Due to the small size of available fundus image datasets, we amalgamated the data used in DoFE to evaluate the performance of SAM. 
% Such an approach serves to alleviate the possibility of incidental experimental results. In the follow-up process, 
Apart from an overall evaluation, we also explored and examined the domain generalization of SAM based on the same domain partitions as DoFE.

% In this benchmark, we utilized a few prompts for OC and OD segmentation to explore SAM's zero-shot performance on enclosed regions \textcolor{red}{what is difference between enclosed and closed mentioned earlier? not annular??} \textcolor{blue}{not pretty sure about difference between those, enclosed should be distinct from the branching structures as vessels of fundus image} of fundus images. 
Note that OC constitutes a whitish, cup-shaped area situated at the center of the OD, which can be approximated as two concentric circles of different radius. Specifically, the fundus images were first cropped to better observe the region of interest (ROI). The positions of two prompts exhibit a high degree of proximity, occasionally coinciding with each other.
% \textcolor{blue}{In terms of choosing appropriate prompts, other choices could be considered, e.g., bounding boxes, as prompts of proximal location are difficult to accurately guide the segmentation of OC and OD.} 
Thus, accurately segmenting between OC and OD is not easy when receiving similar prompts. This partly causes zero-shot SAM to only have half of the accuracy of current SOTAs. We hypothesize that manually prompting SAM (i.e., oracle) may lead to a better accuracy though the zero-shot mode of SAM may still lag behind SOTA by a noticeable margin.

% It is conjectured that the difference in segmentation results between these two types is due to the fact that the area of OC is smaller than OD, and OC is the higher contrast part of the ROI, which may be more common in the training set of SAM from natural scenes.

We also investigated the performance of SAM on cross-domain scenarios following the setting in DoFE \cite{wang2020dofe}. The imbalanced performance observed in Table \ref{fundus} demonstrates that the domain generalization (DG) problem also exists in SAM. SAM achieves the best and worst mean performance on domain 1 and domain 4.
% The major differences between domains result from different imaging devices or scan protocols. These appearance discrepancies would affect the generalization ability of the model, which caused DG issues. 
By inspecting fundus images from four domains, we found that the degree of contrast between OC/OD and background is proportional to the final performance. However, this proportional relationship cannot be observed in DG-optimized algorithms in Table \ref{fundus}. This comparison reveals that SAM mainly focuses on superficial features (image contrast) instead of high-level semantics when zero-shot segmenting OC/OD in fundus images. Figures~\ref{fig:good-samples}B and~\ref{fig:bad-samples}B show some successful and failed fundus segmentation samples, respectively.

\begin{table*}[]
\centering
% \caption{}
\caption{Comparison with SOTA Methods on Fundus Datasets Evaluated using the Dice Similarity Coefficient.}
\label{fundus}
\scalebox{1}{%
\begin{tabular}{@{}cccccccccc@{}}
\toprule
\multirow{2}{*}{Method} & \multicolumn{2}{c}{Domain 1} & \multicolumn{2}{c}{Domain 2} & \multicolumn{2}{c}{Domain 3} & \multicolumn{2}{c}{Domain 4} & Mean \\ \cmidrule(l){2-10} 
      & OC     & OD     & OC     & OD     & OC     & OD     & OC     & OD     & Total           \\ \midrule
U-Net~\cite{ronneberger2015unet} & 0.7703 & 0.9496 & 0.7821 & 0.8969 & 0.8028 & 0.8933 & 0.8474 & 0.9009 & 0.8554          \\
Mixup~\cite{zhang2017mixup} & 0.7332 & 0.9297 & 0.7112 & 0.8678 & 0.8216 & 0.9042 & 0.8623 & 0.9076 & 0.8423          \\
DST~\cite{zhang2019DST}   & 0.7563 & 0.9220 & \textbf{0.8080} & 0.9077 & 0.8432 & \textbf{0.9402} & 0.8624 & 0.9066 & 0.8683          \\
JiGen~\cite{carlucci2019jigen} & 0.8081 & 0.9503 & 0.7946 & 0.9047 & 0.8265 & 0.9194 & 0.8430 & 0.9106 & 0.8697          \\
DoFE~\cite{wang2020dofe}  & \textbf{0.8359} & \textbf{0.9559} & 0.8000 & \textbf{0.9837} & \textbf{0.8666} & 0.9198 & \textbf{0.8704} & \textbf{0.9332} & \textbf{0.8844} \\ \bottomrule
SAM   & 0.5710 & 0.5563 & 0.5200 & 0.3333 & 0.5830 & 0.4157 & 0.3598 & 0.4056 & 0.4609          \\ \bottomrule
\end{tabular}%
}
\end{table*}

\subsubsection{Endoscopic OCT}
For endoscopic OCT, the OCT Rat Colon dataset \cite{park2017broadbandoct} was used, which was captured by an 800-nm ultrahigh-resolution endoscopic SD-OCT system \cite{yuan2022vivo}. The colonic wall is composed of three distinct layers for segmentation, namely colonic mucosa (CM), submucosa (SM), and muscularis externa (ME). A tenth of dataset was randomly sampled in our experiment to evaluate the performance of SAM. Images in this dataset do not contain much color information and complex structures, which is supposed to be easy for segmentation. Nevertheless, the zero-shot segmentation performance of SAM on endoscopic OCT is far inferior to models specifically designed for medical images as shown in Table \ref{rat oct}. Figures \ref{fig:good-samples}G and \ref{fig:bad-samples}G show some segmentation samples of SAM. It can be observed that compared to successful examples of other datasets, the successful examples of endoscopic OCT are actually not perfect, while the failure cases are indeed unsatisfactory.

\begin{table}[]
\centering
\caption{Comparison with SOTA Methods on the OCT Rat Colon Dataset.}
\label{rat oct}
\begin{tabular}{@{}cccc@{}}
\toprule
Method                   & Class     & Dice            & IoU             \\ \midrule
TransUNet~\cite{chen2021transunet}                     & All & \textbf{0.9265} & \textbf{-} \\ 
LiDeOCTNet~\cite{abaxi2023lideoctnet}                     & All & 0.9198 & \textbf{-} \\ \midrule
\multirow{4}{*}{SAM} & Colonic Mucosa  & 0.3491          & 0.2350          \\
                         & Submucosa  & 0.2477          & 0.1485          \\
                         & Muscularis Externa  & 0.2399          & 0.1466          \\
                         & All & 0.2789          & 0.1767          \\ \bottomrule 
\end{tabular}
\end{table}
    
\subsubsection{Ophthalmic OCT} 
Annotated Retinal OCT Images Database (AROI) \cite{melinvsvcak2021aroi} contains 1136 annotated B-scans and associated raw high-resolution images from 24 patients with age-related macular degeneration (AMD). An ophthalmologist annotated three retinal layers and three retinal fluids in each B-scan. The official annotations on B-scan are utilized to classify eight layered structures, including the inner plexiform layer and inner nuclear layer (IPL/INL), as well as regions above internal limiting membrane (ILM) and under bruch's membrane (BM). Pigment epithelial detachment (PED), subretinal fluid and subretinal hyperreflective material (SRF), and intraretinal fluid (IRF) are also involved for segmentation. We did not take the top layer (above ILM) into consideration (i.e., not tested), because it is not a part of internal structure of retina.
% The credibility of predictive accuracy is not entirely dependable, whereby regions that are easily segmented exhibit pronounced precision while non-coherent and small areas yield unreliable segmentation results. 
% As a consequence, when calculating overall accuracy, each partitioned segment is weighted equally, which limits the ability of this measure to accurately assess the predictive capacity of SAM for the eight distinct segments. 

As shown in Table \ref{oct}, the segmentation results for under BM are much higher than other classes, which is also confirmed by the qualitative samples shown in Figure \ref{fig:good-samples}H. This indicates that SAM can accurately segment the structures of clear boundaries and large continuous areas. However, the zero-shot segmentation performance of SAM is extremely unsatisfactory for small holes and elongated layered structures of irregular boundaries (such as RPE-BM), as shown in Figure \ref{fig:bad-samples}H.

% \usepackage{booktabs}
\begin{table}[]
\centering
\caption{Zero-Shot Segmentation Results of SAM for Different Classes from the AROI Dataset.}
\label{oct}
\begin{tabular}{@{}ccc@{}}
\toprule
Class & Dice            & IoU             \\ \midrule
ILM-IPL/INL   & 0.2378          & 0.1527          \\
IPL/INL-RPE   & 0.4499          & 0.3153          \\
RPE-BM   & 0.0688          & 0.0368          \\
under BM   & \textbf{0.8704} & \textbf{0.7806} \\
PED   & 0.1083          & 0.0673          \\
SRF   & 0.1084          & 0.0656          \\
IRF   & 0.0923          & 0.0548          \\ \midrule
Average  & 0.3237          & 0.2506          \\ \bottomrule
\end{tabular}
\end{table}

\subsubsection{CT}  
AMOS \cite{ji2022amos} contains both CT and MRI modalities for abdominal multi-organ segmentation tasks. The entire dataset introduces 500 CT and 100 MRI volume data from 15 organs in abdominal cavity. During pre-processing, we set the CT window range to [$-$991, 362] HU and 15361 slices from the dataset were generated using this standard. Since the corresponding ground truth in the test set has not been released, we used the validation set which contains segmentation labels. Fifteen organs were segmented, including spleen, right kidney, left kidney, gallbladder, esophagus, liver, stomach, aorta, inferior vena cava, pancreas, right adrenal gland, left adrenal gland, duodenum, bladder, prostate and uterus. The above described are also applied to MRI and the experimental results of CT are therefore presented and discussed along with those of MRI in the next section for clarity. 

% \begin{table}[]
% \centering
% \caption{}
% \label{undefined}
% \begin{tabular}{@{}ccc@{}}
% \toprule
% Class   & Mean Dice       & Mean IoU        \\ \midrule
% \#1     & 0.1121          & 0.0612          \\
% \#2     & 0.2687          & 0.2202          \\
% \#3     & 0.3288          & 0.2737          \\
% \#4     & 0.1198          & 0.0916          \\
% \#5     & 0.1442          & 0.1056          \\
% \#6     & 0.3056          & 0.2044          \\
% \#7     & 0.2428          & 0.1587          \\
% \#8     & \textbf{0.3332} & \textbf{0.2748} \\
% \#9     & 0.1429          & 0.1051          \\
% \#10    & 0.1547          & 0.1052          \\
% \#11    & 0.0768          & 0.0628          \\
% \#12    & 0.1114          & 0.0852          \\
% \#13    & 0.1823          & 0.1275          \\
% \#14    & 0.1137          & 0.0622          \\
% \#15    & 0.0404          & 0.0211          \\ \midrule
% Total & 0.2154          & 0.1609          \\ \bottomrule
% \end{tabular}
% \end{table}
\begin{table}[]
\centering
\caption{Zero-Shot Segmentation Results of SAM on Different CT Organ Classes in AMOS Dataset.}
\label{sam on CT}
\begin{tabular}{@{}ccc@{}}
\toprule
Class   & Dice       & IoU        \\ \midrule
\#1     & 0.1616          & 0.1026          \\
\#2     & 0.2723          & 0.2211          \\
\#3     & \textbf{0.3465}          & \textbf{0.2856}          \\
\#4     & 0.0943          & 0.0630          \\
\#5     & 0.1023          & 0.0724          \\
\#6     & 0.3188          & 0.2097          \\
\#7     & 0.2624          & 0.1812          \\
\#8     & 0.2946          & 0.2310 \\
\#9     & 0.1555          & 0.1188          \\
\#10    & 0.1375          & 0.0930          \\
\#11    & 0.0287          & 0.0202          \\
\#12    & 0.0585          & 0.0454          \\
\#13    & 0.1831          & 0.1284          \\
\#14    & 0.1292          & 0.0752          \\
\#15    & 0.0653          & 0.0360          \\ \midrule
Average & 0.2105          & 0.1548          \\ \bottomrule
\end{tabular}
\end{table}

\subsubsection{MRI} 
 The pre-processing method of MRI data is congruent with the one used for CT data mentioned in the preceding section. A total of 3176 images originating from the validation set were obtained and selected for our case study. As the datasets of two modalities are derived from the same dataset and the 3D data were pre-processed in a same way, the average Dice scores of the two modalities are close as shown in Table~\ref{sam on CT} and Table~\ref{sam on MRI}, although the variations exist across individual class segmentation tasks. As shown in Figure \ref{fig:good-samples}C, D, SAM can accurately segment two-dimensional organs in CT and MRI images. The successful segmentation manifests the precise localization and clear demarcation, and a striking resemblance to the actual organ structures in terms of visual appearance. However, the failure cases shown in \mbox{Figure \ref{fig:bad-samples}C, D} are often from the segmentation of small organs, in which SAM exhibits a tendency to segment not only the target organ itself but also the neighboring tissues.
% \begin{table}[]
% \centering
% \caption{}
% \label{undefined}
% \begin{tabular}{@{}ccc@{}}
% \toprule
% Class   & Mean Dice & Mean IoU \\ \midrule
% \#1     & 0.4252    & 0.3195   \\
% \#2     & 0.5556    & 0.4781   \\
% \#3     & 0.4991    & 0.3981   \\
% \#4     & 0.0331    & 0.0172   \\
% \#5     & 0.0431    & 0.0246   \\
% \#6     & 0.5206    & 0.4203   \\
% \#7     & 0.1037    & 0.0556   \\
% \#8     & 0.3877    & 0.3177   \\
% \#9     & 0.1111    & 0.0816   \\
% \#10    & 0.1591    & 0.1069   \\
% \#11    & 0.0080    & 0.0040   \\
% \#12    & 0.1897    & 0.1500   \\
% \#13    & 0.1887    & 0.1341   \\ \midrule
% Total & 0.2271    & 0.1731   \\ \bottomrule
% \end{tabular}
% \end{table}
\begin{table}[]
\centering
\caption{Zero-Shot Segmentation Results of SAM on Different MRI Organ Classes in AMOS Dataset.}
\label{sam on MRI}
\begin{tabular}{@{}ccc@{}}
\toprule
Class   & Dice & IoU \\ \midrule
\#1     & 0.2571    & 0.1785   \\
\#2     & \textbf{0.4622}    & \textbf{0.3866}   \\
\#3     & 0.4434    & 0.3656   \\
\#4     & 0.1437    & 0.1026   \\
\#5     & 0.0352    & 0.0207   \\
\#6     & 0.4480    & 0.3311   \\
\#7     & 0.1475    & 0.0912   \\
\#8     & 0.2843    & 0.2285   \\
\#9     & 0.0691    & 0.0452   \\
\#10    & 0.1175    & 0.0754   \\
\#11    & 0.0054    & 0.0027   \\
\#12    & 0.0243    & 0.0177   \\
\#13    & 0.0989    & 0.0642   \\ \midrule
Average & 0.2264    & 0.1723   \\ \bottomrule
\end{tabular}
\end{table}

\subsubsection{X-Ray}    
 The dataset (704 labelled images) used for testing SAM on X-ray is compiled from the validation and training sets of \cite{candemir2013lungxray,jaeger2013automaticXRay}. As shown in Figure \ref{fig:good-samples}F, the segmentation performance demonstrated by SAM in successful cases is impressive and the segmentation results perfectly match with the ground truth. However, the failure cases shown in \mbox{Figure \ref{fig:bad-samples}F} reveal that the zero-shot chest X-ray segmentation of SAM is not always consistent. Table \ref{x-ray} shows the quantitative results of zero-shot X-ray image segmentation.

% \usepackage{booktabs}
\begin{table}[]
\centering
\caption{Quantitative Results of Zero-Shot SAM on an X-ray Dataset.}
\label{x-ray}
\begin{tabular}{@{}ccc@{}}
\toprule
Method  & Dice   & IoU    \\ \midrule
SAM & 0.6509 & 0.5136 \\ \bottomrule
\end{tabular}
\end{table}

\subsubsection{Endoscopic Images}   
We further tested SAM on the MICCAI 2017 Robotic Instrument Segmentation dataset \cite{shvets2018automaticinstrument}, which contains 2040 stereo camera images acquired from a da Vinci Xi surgical system on three types of segmentation task (binary, parts and instruments segmentation). The images are essentially captured by RGB cameras. As a result, the segmentation outcomes shown in Figure \ref{fig:good-samples}E are impressive when appropriate prompts are given. As we place the prompt at the center of each individual object, a three-class segmentation of parts is the most suitable choice for testing in order to avoid any irregular connections and interference between different parts compared with binary segmentation and instrument segmentation. The performance of SAM was tested using data from the official testing set, and the results were compared against those provided in the official report. To better visualize, segmentation results of different categories have been plotted onto the same graph for comparison, which are shown in both Figures \ref{fig:good-samples}E and \ref{fig:bad-samples}E. {It should be noted that even though some examples exhibit visually satisfactory results, they are likely due to the predicted masks partially lying within other masks.} From a quantitative perspective as shown in Table \ref{medrobo}, SAM has already surpassed the segmentation performance of LinkNet-34 \cite{shvets2018automaticinstrument,chaurasia2017linknet} by only receiving prompts. However, it is still not comparable to the current SOTA and there is a 30\% relative gap between SAM and SOTA methods.

% \usepackage{booktabs}
% \usepackage{multirow}
\begin{table}[]
\centering
\caption{Comparison with SOTA Methods on the MICCAI 2017 Robotic Instrument Segmentation Dataset.}
\label{medrobo}
\begin{tabular}{@{}cccc@{}}
\toprule
Method                   & Class   & Dice            & IoU             \\ \midrule
UNet~\cite{ronneberger2015unet}                     & All    & 0.6075          & 0.4841          \\ 
TernausNet-16~\cite{shvets2018automaticinstrument}            & All    & \textbf{0.7597} & \textbf{0.6550} \\ 
TernausNet-11~\cite{shvets2018automaticinstrument}            & All    & 0.7425          & 0.6223          \\ 
LinkNet-34~\cite{shvets2018automaticinstrument}               & All    & 0.4126          & 0.3455          \\ \midrule
\multirow{4}{*}{SAM} & Clasper & 0.4296          & 0.3054          \\
                         & Wrist   & 0.5076          & 0.3674          \\
                         & Shaft   & 0.7189          & 0.5076          \\
                         & All    & 0.5512          & 0.4227          \\ \bottomrule
\end{tabular}
\end{table}

\subsubsection{Retinal Vessel in Fundus Images}
    
\begin{figure}[htbp] 
	
	\centering
	
	\includegraphics[width=\linewidth,scale=0.5]{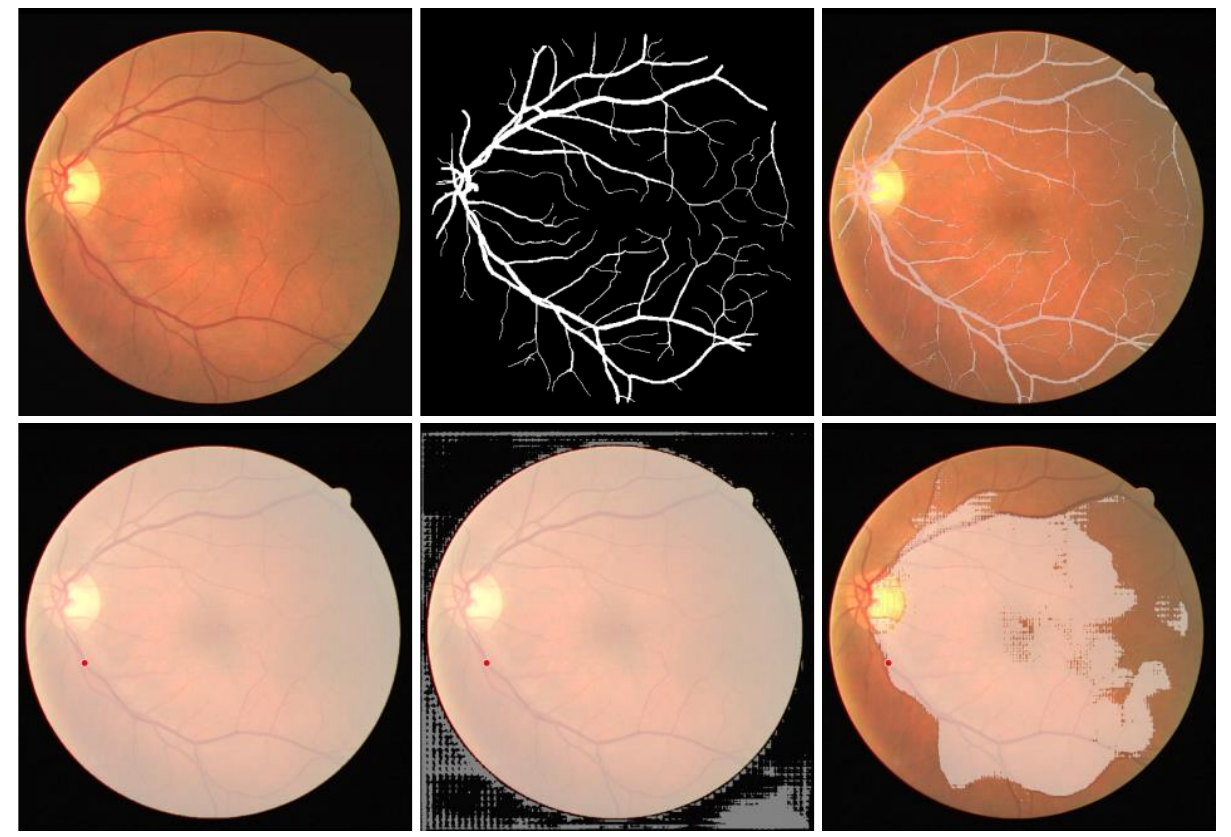}
	\caption{A failure sample of SAM on segmenting retinal vessels. The first row from left to right is: the initial input image, ground truth mask, and the input image superimposed with the ground truth mask. The second row from left to right shows three SAM segmented images with the score of 1.007, 0.993, and 0.673, respectively.}
	\label{fail-sample}
	
\end{figure}

\begin{figure}[htbp] 
	
	\centering
	
	\includegraphics[width=\linewidth,scale=0.5]{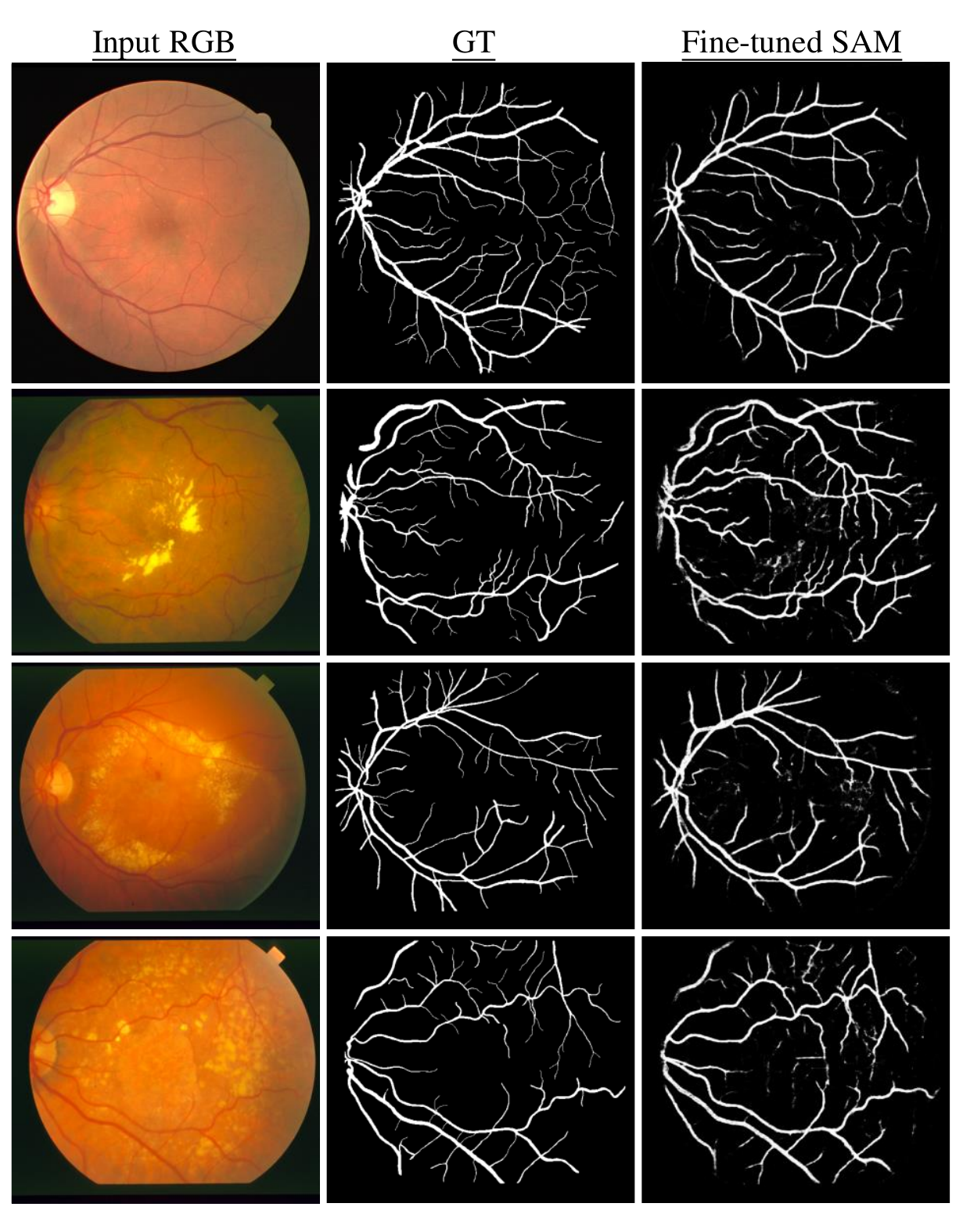}
	\caption{Segmentation samples of SAM fine-tuned on retinal vessels. Each row from left to right is the initial input image, ground truth mask and prediction of fine-tuned SAM. The column from top to bottom shows retinal images from four different datasets.}
	\label{finetune-sample}
	
\end{figure}

We also used SAM to segment vessels in fundus images. However, the experiments revealed that zero-shot SAM is not able to accurately segment retinal blood vessels, despite the \textit{manual} provision of additional prompts focused on areas where visibility of vessels is pronounced. At the current stage, we conjectured that the segmentation of continuously branching structures, such as blood vessels in medical images and tree branches in nature images, presents a challenge for SAM. As shown in Figure \ref{fail-sample}, it is found that SAM encounters difficulties in accurately identifying vessels as distinct segmentable objects.
Therefore, we further conducted fine-tuning of SAM to examine its potential of improving segmentation results. To fine tune SAM (SAM ViT-B in this experiment), we utilized SAM adapter \cite{chen2023samadapter}, a task-specific fine-tuning method proposed for SAM. Specifically, 20 image-mask pairs were selected from the Digital Retinal Images for Vessel Extraction (DRIVE) dataset \cite{staal2004ridgeDRIVE} to fine-tune SAM's mask decoder. As the fine-tuning was fully supervised, there was no prompt provided and the ground truth masks were used to supervise the training. The entire fine-tuning process was trained with the AdamW \cite{loshchilov2017adamw} optimizer. The learning rate was set to {$2 \times 10^{-4}$} %MDPI: please change the terms into scientific notations, e.g., “8 × 10³”, not “8E3”.
with a total of 20 training epochs. The batch size was one with the input image size of 1024 $\times$ 1024 pixels. A total of four datasets were utilized for the quantitative assessment. Three of them are the official test sets from their respective datasets. The official training set from the STARE dataset \cite{hoover2000STARE} was also employed for testing as its own test set is lacking. As shown in Table \ref{quatitive result of fine-turn}, the fine-tuned SAM demonstrated substantial improvements in both the Dice and IoU metrics. Noticeably, the Dice scores increased by at least 200 percent relative to its zero-shot counterpart where vessels were segmented by manually providing the prompt. Furthermore, the improvements of IoU were even more pronounced. In addition, as illustrated in Figure \ref{finetune-sample}, the segmentation results of SAM after fine-tuning almost perfectly match with ground truth in some cases and only some tiny parts at the terminal ends of vessels are missing, demonstrating the potential of SAM for precise medical image segmentation after a domain-specific fine-tuning.    

\begin{table*}[]
\centering
\caption{Quantitative Results of Fine-tuned SAM on Retinal Vessel Segmentation Datasets.}
\label{quatitive result of fine-turn}
\resizebox{\linewidth}{!}{
\begin{tabular}{@{}ccccccc@{}}
\toprule
Dataset  & Dice (Manual) & Dice (Fine-tuned) & Dice Relative Increase & IoU (Manual) & IoU (Fine-tuned) & IoU Relative Increase \\ \midrule
DRIVE~\cite{staal2004ridgeDRIVE}    & 0.2267       & 0.7733         & 241\%                  & 0.1281      & 0.6304        & 392\%                 \\
CHASEDB1~\cite{fraz2012CHASEDB1} & 0.1685       & 0.7118         & 322\%                  & 0.0921      & 0.5526        & 500\%                 \\
HRF~\cite{odstrcilik2013HRF}      & 0.1701       & 0.6776         & 298\%                  & 0.0931      & 0.5124        & 450\%                 \\
STARE~\cite{hoover2000STARE}    & 0.1902       & 0.7694         & 304\%                  & 0.1057      & 0.6252        & 491\%                 \\ \bottomrule
\end{tabular}}
\end{table*}

% \end{enumerate}
\section{Discussion}\label{sec:discussion}

Through extensive experiments, it is found that SAM is unable to outperform models specially designed for medical imaging by simply using its zero-shot feature. It is worth noting that SAM is trained primarily on natural images and has limited access to diverse medical modalities containing pathological manifestations. As such, the medical segmentation results obtained using SAM's zero-shot capability are actually decent and sometimes considered impressive as shown in Figure \ref{fig:good-samples}. Nevertheless, given prompts in the same way, the performance of SAM varies significantly across different medical modalities. In endoscopy and dermoscopy, SAM demonstrated better performance compared to other medical modalities. In segmenting the skin lesion, if there are no clear boundaries of the lesion or if the skin itself has pigment deposition and erythematosus, the segmentation outcome of SAM is often unsatisfactory. This phenomenon resembles the concealed patterns observed in natural images, which have been previously shown to be a challenge for SAM~\cite{tang2023can,ji2023samcan}. Although the retinal fundus images are also RGB images, segmenting structural targets within the internal structure of retina is not encountered in natural scenes. This has resulted in suboptimal segmentation of optic disc and complete failure in retinal vessel segmentation of SAM using zero-shot segmentation.

Despite the SA-1B dataset used for training SAM contains 1B masks and 11M images, the multiple medical domains we tested in our study are entirely unseen domains for SAM. In particular, 3D imaging predominates as a critical medical methodology, such as MRI, CT, and OCT. The 2D slices employed for analysis are the unique aspects of medical imaging and cannot be found in natural domains. For instance, the features of OCT B-scan are layered structures stretching along the entire image width, instead of closed regions. Algorithms developed specifically for the prominent features of OCT have demonstrated excellent performance \cite{viedma2022deepoct,abaxi2023lideoctnet}. However, SAM is unable to discriminate the tissue layers in OCT images without any prior knowledge. In addition to the presence of domain differences between medical and natural images, it has been observed in Table \ref{fundus} that SAM exhibits significantly imbalanced segmentation accuracy when encountering different domain images under the same category. 

% All mentions above serve as an inspiration to leverage the robust feature extraction capabilities of the SAM at hand to specific medical modalities and further healthcare applications. \textcolor{red}{see my comments}

To evaluate the capabilities of SAM under zero-shot settings, experiments in this study used a single prompt selection method and used the center point of the ground truth mask as the prompt for each sample. Although this approach did not fully harness the potential of SAM, it suffices to highlight the limitations of SAM in medical imaging. Recently, SAM-Adapter \cite{chen2023samadapter} has been proposed for tackling complex scenario segmentation on natural image datasets including but not limited to camouflaged targets and defocus blur objects, which has shown better results than SAM and other task-specific approaches. Medical images can be regarded as a distinct category of rare scenes. Consequently, it is very likely that natural image-based large models after fine-tuning may yield excellent performance on medical imaging modalities as revealed by our preliminary results of SAM fine-tuned on retinal vessel data. Meanwhile, different prompt engineering techniques can be further explored in the future. Furthermore, it is worth investigating if training a large medical vision model from scratch using only medical data can lead to a better performance than continual-training/fine-tuning a large vision model using medical images, which has been previously pretrained on a large volume of natural image data. In addition, during the development phase of large medical AI models, it is recommended to prioritize a focus on diagnostic information and invariant features present in medical images. This can potentially mitigate issues related to domain transfer, thus enhancing the overall performance and interpretability of a large AI model.

{The limitations of our testing method are also worth noting, and future works are encouraged to further explore SAM on medical segmentation. Firstly, we only used one of the three prompts provided in the official setting. For some cases, the centroid might not be inside the region of interest, other segment modes of SAM should be taken into consideration, such as the bounding box mode and automatically segment everything mode, which may provide a more comprehensive evaluation of its segmentation ability. Another limitation is that we only performed fine-tuning on retinal vessel datasets. Although the results indicate that the fine-tuned model demonstrates excellent segmentation quality compared to its zero-shot counterpart, we believe experiments on a wider range of datasets are needed to examine the segmentation performance of fine-tuned SAM, which could provide a more holistic view of its pros and cons on different medical modalities.}

% while fine-tuning the encoder of existing vision foundation models, the enormous number of parameters presents a significant barrier for groups with limited computational resources. Therefore, incorporating prompts into the encoder can enhance the model's perception of new datasets, enabling it to capture more specific features with greater detail. 
% Simultaneously, during the process of fine-tuning a model, it is worth investigating whether retaining the original knowledge of the model is beneficial. 
% This stems from an unresolved inquiry regarding whether pre-training on features extracted from natural images necessarily confers advantages to tasks involving medical images. Thus, implementing a suitable fine-tuning or additional strategy that promotes invariant feature learning could constitute a vital step towards incorporating vision foundation models into the analysis of medical images.

{Nevertheless, our work is the first research work that focuses on evaluating and analyzing the performance of a recently developed large AI model, i.e., SAM on a wide range of medical image segmentation tasks both qualitatively and quantitatively, with detailed comparisons between SAM and baselines. The findings from our research help identify where SAM works and how it can be fine-tuned to provide better performance on medical imaging and applications. Our results may also help guide the future development of SAM and other medical generalist AI models on domain-specific tasks. 
Medical image segmentation is an important and challenging task. If the advanced large models such as SAM become highly accurate and robust on medical image segmentation, either in a zero-shot fashion or after fine-tuning, they may bring in far-reaching impact and help transform and improve medical diagnostics, treatment planning, and other healthcare applications that depend on medical imaging. This could eventually lead to great benefits to both clinicians and patients. 
In summary, our study lays an important groundwork for developing and applying large AI models on medical image segmentation. With continued progress, such models could positively impact healthcare by assisting with and improving critical tasks in medical diagnostics. Our current work not only highlights the potential significance and societal benefits of this line of research, but also identifies the limitations and the needs of further research before the achievement of substantial real-world impacts. 
}
\section{Conclusions}\label{sec:conclusions}
This work presents a benchmark study of SAM on a wide range of zero-shot medical image segmentation tasks. Through comprehensive experiments, we identify the challenges that SAM currently encounters in this context. Importantly, our analysis is performed on a standardized set of prompts devoid of any prior medical knowledge, covering a diverse range of imaging modalities, such as dermoscope, fundus, CT, MRI, endoscope, X-Ray, endoscopic OCT and ophthalmic OCT. The provision of precise and interactive prompts, the use of specialized feature extraction methodologies tailored for medical images, and the well-designed fine-tuning strategies of large vision models originally trained on natural images can be explored in future works. Given the unique challenges associated with medical imaging, these aspects are critical for ensuring the optimal performance of generalist models in this domain. Additionally, it is crucial that large medical vision models possess cross-domain generalizability, akin to that exhibited by physicians. This is important to avoid any negative impact on diagnostic accuracy resulting from the use of new equipment and protocols. Overall, our findings highlight the needs for medical foundation models with careful consideration given to the specific challenges posed by this complex and rapidly evolving field.

%%%%%%%%% REFERENCES
{\small
\bibliographystyle{ieee_fullname}
% \bibliography{reference}

\begin{thebibliography}{10}\itemsep=-1pt

\bibitem{abaxi2023lideoctnet}
Sai Mu~Dalike Abaxi, Mehmood Nawaz, Peilun Shi, Hao Wei, Syeda~Aimen Abbasi,
  and Wu Yuan.
\newblock Lideoctnet: A lightweight oct-aware framework for segmentation of
  irregularly layered tissue structures.
\newblock 2023.

\bibitem{allan20192017robodata}
Max Allan, Alex Shvets, Thomas Kurmann, Zichen Zhang, Rahul Duggal, Yun-Hsuan
  Su, Nicola Rieke, Iro Laina, Niveditha Kalavakonda, Sebastian Bodenstedt,
  et~al.
\newblock 2017 robotic instrument segmentation challenge.
\newblock {\em arXiv preprint arXiv:1902.06426}, 2019.

\bibitem{benvcevic2021skinsota}
Marin Ben{\v{c}}evi{\'c}, Irena Gali{\'c}, Marija Habijan, and Danilo Babin.
\newblock Training on polar image transformations improves biomedical image
  segmentation.
\newblock {\em IEEE Access}, 9:133365--133375, 2021.

\bibitem{bommasani2021opportunities}
Rishi Bommasani, Drew~A Hudson, Ehsan Adeli, Russ Altman, Simran Arora, Sydney
  von Arx, Michael~S Bernstein, Jeannette Bohg, Antoine Bosselut, Emma
  Brunskill, et~al.
\newblock On the opportunities and risks of foundation models.
\newblock {\em arXiv preprint arXiv:2108.07258}, 2021.

\bibitem{candemir2013lungxray}
Sema Candemir, Stefan Jaeger, Kannappan Palaniappan, Jonathan~P Musco, Rahul~K
  Singh, Zhiyun Xue, Alexandros Karargyris, Sameer Antani, George Thoma, and
  Clement~J McDonald.
\newblock Lung segmentation in chest radiographs using anatomical atlases with
  nonrigid registration.
\newblock {\em IEEE transactions on medical imaging}, 33(2):577--590, 2013.

\bibitem{carlucci2019jigen}
Fabio~M Carlucci, Antonio D'Innocente, Silvia Bucci, Barbara Caputo, and
  Tatiana Tommasi.
\newblock Domain generalization by solving jigsaw puzzles.
\newblock In {\em Proceedings of the IEEE/CVF Conference on Computer Vision and
  Pattern Recognition}, pages 2229--2238, 2019.

\bibitem{chaurasia2017linknet}
Abhishek Chaurasia and Eugenio Culurciello.
\newblock Linknet: Exploiting encoder representations for efficient semantic
  segmentation.
\newblock In {\em 2017 IEEE visual communications and image processing (VCIP)},
  pages 1--4. IEEE, 2017.

\bibitem{chen2021transunet}
Jieneng Chen, Yongyi Lu, Qihang Yu, Xiangde Luo, Ehsan Adeli, Yan Wang, Le Lu,
  Alan~L Yuille, and Yuyin Zhou.
\newblock Transunet: Transformers make strong encoders for medical image
  segmentation.
\newblock {\em arXiv preprint arXiv:2102.04306}, 2021.

\bibitem{chen2023samadapter}
Tianrun Chen, Lanyun Zhu, Chaotao Ding, Runlong Cao, Shangzhan Zhang, Yan Wang,
  Zejian Li, Lingyun Sun, Papa Mao, and Ying Zang.
\newblock Sam fails to segment anything? -- sam-adapter: Adapting sam in
  underperformed scenes: Camouflage, shadow, and more, 2023.

\bibitem{codella2019skin}
Noel Codella, Veronica Rotemberg, Philipp Tschandl, M~Emre Celebi, Stephen
  Dusza, David Gutman, Brian Helba, Aadi Kalloo, Konstantinos Liopyris, Michael
  Marchetti, et~al.
\newblock Skin lesion analysis toward melanoma detection 2018: A challenge
  hosted by the international skin imaging collaboration (isic).
\newblock {\em arXiv preprint arXiv:1902.03368}, 2019.

\bibitem{deng2023segment}
Ruining Deng, Can Cui, Quan Liu, Tianyuan Yao, Lucas~W Remedios, Shunxing Bao,
  Bennett~A Landman, Lee~E Wheless, Lori~A Coburn, Keith~T Wilson, et~al.
\newblock Segment anything model (sam) for digital pathology: Assess zero-shot
  segmentation on whole slide imaging.
\newblock {\em arXiv preprint arXiv:2304.04155}, 2023.

\bibitem{dong2021polyp}
Bo Dong, Wenhai Wang, Deng-Ping Fan, Jinpeng Li, Huazhu Fu, and Ling Shao.
\newblock Polyp-pvt: Polyp segmentation with pyramid vision transformers.
\newblock {\em arXiv preprint arXiv:2108.06932}, 2021.

\bibitem{fan2020pranet}
Deng-Ping Fan, Ge-Peng Ji, Tao Zhou, Geng Chen, Huazhu Fu, Jianbing Shen, and
  Ling Shao.
\newblock Pranet: Parallel reverse attention network for polyp segmentation.
\newblock In {\em Medical Image Computing and Computer Assisted
  Intervention--MICCAI 2020: 23rd International Conference, Lima, Peru, October
  4--8, 2020, Proceedings, Part VI 23}, pages 263--273. Springer, 2020.

\bibitem{fraz2012CHASEDB1}
Muhammad~Moazam Fraz, Paolo Remagnino, Andreas Hoppe, Bunyarit Uyyanonvara,
  Alicja~R Rudnicka, Christopher~G Owen, and Sarah~A Barman.
\newblock An ensemble classification-based approach applied to retinal blood
  vessel segmentation.
\newblock {\em IEEE Transactions on Biomedical Engineering}, 59(9):2538--2548,
  2012.

\bibitem{fumero2011rim}
Francisco Fumero, Silvia Alay{\'o}n, Jos{\'e}~L Sanchez, Jose Sigut, and M
  Gonzalez-Hernandez.
\newblock Rim-one: An open retinal image database for optic nerve evaluation.
\newblock In {\em 2011 24th international symposium on computer-based medical
  systems (CBMS)}, pages 1--6. IEEE, 2011.

\bibitem{hoover2000STARE}
AD Hoover, Valentina Kouznetsova, and Michael Goldbaum.
\newblock Locating blood vessels in retinal images by piecewise threshold
  probing of a matched filter response.
\newblock {\em IEEE Transactions on Medical imaging}, 19(3):203--210, 2000.

\bibitem{jaeger2013automaticXRay}
Stefan Jaeger, Alexandros Karargyris, Sema Candemir, Les Folio, Jenifer
  Siegelman, Fiona Callaghan, Zhiyun Xue, Kannappan Palaniappan, Rahul~K Singh,
  Sameer Antani, et~al.
\newblock Automatic tuberculosis screening using chest radiographs.
\newblock {\em IEEE transactions on medical imaging}, 33(2):233--245, 2013.

\bibitem{jha2020doubleu}
Debesh Jha, Michael~A Riegler, Dag Johansen, P{\aa}l Halvorsen, and
  H{\aa}vard~D Johansen.
\newblock Doubleu-net: A deep convolutional neural network for medical image
  segmentation.
\newblock In {\em 2020 IEEE 33rd International symposium on computer-based
  medical systems (CBMS)}, pages 558--564. IEEE, 2020.

\bibitem{ji2023samcan}
Ge-Peng Ji, Deng-Ping Fan, Peng Xu, Ming-Ming Cheng, Bowen Zhou, and Luc
  Van~Gool.
\newblock Sam struggles in concealed scenes--empirical study on" segment
  anything".
\newblock {\em arXiv preprint arXiv:2304.06022}, 2023.

\bibitem{ji2023segment}
Wei Ji, Jingjing Li, Qi Bi, Wenbo Li, and Li Cheng.
\newblock Segment anything is not always perfect: An investigation of sam on
  different real-world applications.
\newblock {\em arXiv preprint arXiv:2304.05750}, 2023.

\bibitem{ji2022amos}
Yuanfeng Ji, Haotian Bai, Jie Yang, Chongjian Ge, Ye Zhu, Ruimao Zhang, Zhen
  Li, Lingyan Zhang, Wanling Ma, Xiang Wan, et~al.
\newblock Amos: A large-scale abdominal multi-organ benchmark for versatile
  medical image segmentation.
\newblock {\em arXiv preprint arXiv:2206.08023}, 2022.

\bibitem{kirillov2023segment}
Alexander Kirillov, Eric Mintun, Nikhila Ravi, Hanzi Mao, Chloe Rolland, Laura
  Gustafson, Tete Xiao, Spencer Whitehead, Alexander~C Berg, Wan-Yen Lo, et~al.
\newblock Segment anything.
\newblock {\em arXiv preprint arXiv:2304.02643}, 2023.

\bibitem{loshchilov2017adamw}
Ilya Loshchilov and Frank Hutter.
\newblock Decoupled weight decay regularization.
\newblock {\em arXiv preprint arXiv:1711.05101}, 2017.

\bibitem{lou2022caranet}
Ange Lou, Shuyue Guan, Hanseok Ko, and Murray~H Loew.
\newblock Caranet: context axial reverse attention network for segmentation of
  small medical objects.
\newblock In {\em Medical Imaging 2022: Image Processing}, volume 12032, pages
  81--92. SPIE, 2022.

\bibitem{mattjie2023exploring}
Christian Mattjie, Luis~Vinicius de Moura, Rafaela~Cappelari Ravazio,
  Lucas~Silveira Kupssinsk{\"u}, Ot{\'a}vio Parraga, Marcelo~Mussi Delucis, and
  Rodrigo~Coelho Barros.
\newblock Exploring the zero-shot capabilities of the segment anything model
  (sam) in 2d medical imaging: A comprehensive evaluation and practical
  guideline.
\newblock {\em arXiv preprint arXiv:2305.00109}, 2023.

\bibitem{melinvsvcak2021annotatedAROI}
Martina Melin{\v{s}}{\v{c}}ak, Marin Radmilovi{\'c}, Zoran Vatavuk, and Sven
  Lon{\v{c}}ari{\'c}.
\newblock Annotated retinal optical coherence tomography images (aroi) database
  for joint retinal layer and fluid segmentation.
\newblock {\em Automatika}, 62(3-4):375--385, 2021.

\bibitem{melinvsvcak2021aroi}
M Melin{\v{s}}{\v{c}}ak, M Radmilovi{\v{c}}, Zoran Vatavuk, and S
  Lon{\v{c}}ari{\'c}.
\newblock Aroi: Annotated retinal oct images database.
\newblock In {\em 2021 44th International Convention on Information,
  Communication and Electronic Technology (MIPRO)}, pages 371--376. IEEE, 2021.

\bibitem{odstrcilik2013HRF}
Jan Odstrcilik, Radim Kolar, Attila Budai, Joachim Hornegger, Jiri Jan, Jiri
  Gazarek, Tomas Kubena, Pavel Cernosek, Ondrej Svoboda, and Elli Angelopoulou.
\newblock Retinal vessel segmentation by improved matched filtering: evaluation
  on a new high-resolution fundus image database.
\newblock {\em IET Image Processing}, 7(4):373--383, 2013.

\bibitem{orlando2020refuge}
Jos{\'e}~Ignacio Orlando, Huazhu Fu, Jo{\~a}o~Barbosa Breda, Karel Van~Keer,
  Deepti~R Bathula, Andr{\'e}s Diaz-Pinto, Ruogu Fang, Pheng-Ann Heng, Jeyoung
  Kim, JoonHo Lee, et~al.
\newblock Refuge challenge: A unified framework for evaluating automated
  methods for glaucoma assessment from fundus photographs.
\newblock {\em Medical image analysis}, 59:101570, 2020.

\bibitem{park2017broadbandoct}
Hyeon-Cheol Park, Jessica Mavadia-Shukla, Wu Yuan, Milad Alemohammad, and
  Xingde Li.
\newblock Broadband rotary joint for high-speed ultrahigh-resolution endoscopic
  oct imaging at 800 nm.
\newblock {\em Optics letters}, 42(23):4978--4981, 2017.

\bibitem{qiu2023large}
Jianing Qiu, Lin Li, Jiankai Sun, Jiachuan Peng, Peilun Shi, Ruiyang Zhang,
  Yinzhao Dong, Kyle Lam, Frank P-W Lo, Bo Xiao, Wu Yuan, Dong Xu, and Benny
  Lo.
\newblock Large ai models in health informatics: Applications, challenges, and
  the future.
\newblock {\em arXiv preprint arXiv:2303.11568}, 2023.

\bibitem{ronneberger2015unet}
Olaf Ronneberger, Philipp Fischer, and Thomas Brox.
\newblock U-net: Convolutional networks for biomedical image segmentation.
\newblock In {\em Medical Image Computing and Computer-Assisted
  Intervention--MICCAI 2015: 18th International Conference, Munich, Germany,
  October 5-9, 2015, Proceedings, Part III 18}, pages 234--241. Springer, 2015.

\bibitem{shvets2018automaticinstrument}
Alexey~A Shvets, Alexander Rakhlin, Alexandr~A Kalinin, and Vladimir~I
  Iglovikov.
\newblock Automatic instrument segmentation in robot-assisted surgery using
  deep learning.
\newblock In {\em 2018 17th IEEE International Conference on Machine Learning
  and Applications (ICMLA)}, pages 624--628, 2018.

\bibitem{sivaswamy2015comprehensiveDrishti-GS}
Jayanthi Sivaswamy, S Krishnadas, Arunava Chakravarty, G Joshi, A~Syed Tabish,
  et~al.
\newblock A comprehensive retinal image dataset for the assessment of glaucoma
  from the optic nerve head analysis.
\newblock {\em JSM Biomedical Imaging Data Papers}, 2(1):1004, 2015.

\bibitem{staal2004ridgeDRIVE}
Joes Staal, Michael~D Abr{\`a}moff, Meindert Niemeijer, Max~A Viergever, and
  Bram Van~Ginneken.
\newblock Ridge-based vessel segmentation in color images of the retina.
\newblock {\em IEEE transactions on medical imaging}, 23(4):501--509, 2004.

\bibitem{tang2022duat}
Feilong Tang, Qiming Huang, Jinfeng Wang, Xianxu Hou, Jionglong Su, and Jingxin
  Liu.
\newblock Duat: Dual-aggregation transformer network for medical image
  segmentation.
\newblock {\em arXiv preprint arXiv:2212.11677}, 2022.

\bibitem{tang2023can}
Lv Tang, Haoke Xiao, and Bo Li.
\newblock Can sam segment anything? when sam meets camouflaged object
  detection.
\newblock {\em arXiv preprint arXiv:2304.04709}, 2023.

\bibitem{tschandl2018ham10000skin}
Philipp Tschandl, Cliff Rosendahl, and Harald Kittler.
\newblock The ham10000 dataset, a large collection of multi-source
  dermatoscopic images of common pigmented skin lesions.
\newblock {\em Scientific data}, 5(1):1--9, 2018.

\bibitem{viedma2022deepoct}
Ignacio~A Viedma, David Alonso-Caneiro, Scott~A Read, and Michael~J Collins.
\newblock Deep learning in retinal optical coherence tomography (oct): A
  comprehensive survey.
\newblock {\em Neurocomputing}, 2022.

\bibitem{wang2020dofe}
Shujun Wang, Lequan Yu, Kang Li, Xin Yang, Chi-Wing Fu, and Pheng-Ann Heng.
\newblock Dofe: Domain-oriented feature embedding for generalizable fundus
  image segmentation on unseen datasets.
\newblock {\em IEEE Transactions on Medical Imaging}, 2020.

\bibitem{yuan2022vivo}
Wu Yuan, Yan Feng, Defu Chen, Payam Gharibani, Jiande~DZ Chen, Huimin Yu, and
  Xingde Li.
\newblock In vivo assessment of inflammatory bowel disease in rats with
  ultrahigh-resolution colonoscopic oct.
\newblock {\em Biomedical optics express}, 13(4):2091--2102, 2022.

\bibitem{zhang2017mixup}
Hongyi Zhang, Moustapha Cisse, Yann~N Dauphin, and David Lopez-Paz.
\newblock mixup: Beyond empirical risk minimization.
\newblock {\em arXiv preprint arXiv:1710.09412}, 2017.

\bibitem{zhang2019DST}
Ling Zhang, Xiaosong Wang, Dong Yang, Thomas Sanford, Stephanie Harmon, Baris
  Turkbey, Holger Roth, Andriy Myronenko, Daguang Xu, and Ziyue Xu.
\newblock When unseen domain generalization is unnecessary? rethinking data
  augmentation.
\newblock {\em arXiv preprint arXiv:1906.03347}, 2019.

\bibitem{zhang2021transfuse}
Yundong Zhang, Huiye Liu, and Qiang Hu.
\newblock Transfuse: Fusing transformers and cnns for medical image
  segmentation.
\newblock In {\em Medical Image Computing and Computer Assisted
  Intervention--MICCAI 2021: 24th International Conference, Strasbourg, France,
  September 27--October 1, 2021, Proceedings, Part I 24}, pages 14--24.
  Springer, 2021.

\bibitem{zhou2018unet++}
Zongwei Zhou, Md~Mahfuzur Rahman~Siddiquee, Nima Tajbakhsh, and Jianming Liang.
\newblock Unet++: A nested u-net architecture for medical image segmentation.
\newblock In {\em Deep Learning in Medical Image Analysis and Multimodal
  Learning for Clinical Decision Support: 4th International Workshop, DLMIA
  2018, and 8th International Workshop, ML-CDS 2018, Held in Conjunction with
  MICCAI 2018, Granada, Spain, September 20, 2018, Proceedings 4}, pages 3--11.
  Springer, 2018.

\end{thebibliography}

}

\end{document}